\documentclass{article}

\usepackage{microtype}
\usepackage{graphicx}
\usepackage{subfigure}
\usepackage{booktabs} 

\usepackage{hyperref} 

\usepackage[accepted]{icml2020_MODIFIED} 

\icmltitlerunning{Translating Diffusion, Wavelets, and      
                  Regularisation into Residual Networks}

\usepackage{pgfplots}
\usepackage{makecell}
\usepackage{tikz}
\usepackage{amsmath, amssymb}
\usepackage{pdflscape}
\usepackage{afterpage}
\usepackage{bm}
\usepackage{diagbox}
\usepackage{bigints}
\usepackage{amsthm}

\usetikzlibrary{matrix,calc,shapes}

\pgfplotsset{%
  basestyle/.style={%
    samples=300,
    domain=-10:10, 
    axis lines=middle,
    xmin = -2,
    xmax = 2,
    ymin = -2,
    ymax = 2,
    xtick = {1},
    ytick = {1},
    tick style = {thick, black},
    major tick length = 7,
    x label style={at={(axis description cs:1.0,0.47)},anchor=north},
    y label style={at={(axis description cs:0.4,1.05)},anchor=north},
    width=\axisdefaultwidth,
    height=0.8*\axisdefaultwidth,
    font = \LARGE},
  homstyle/.style = {basestyle},
  charbstyle/.style = {basestyle,
                       xtick = {\la},
                       ytick = {1},
                       xticklabels = {$\lambda$},
                       yticklabels = {$1$}},
  tvstyle/.style = {basestyle},
  pmistyle/.style = {charbstyle},
  pmiistyle/.style = {charbstyle},
  fabstyle/.style = {basestyle,
                     xtick = {\la, \lb},
                     ytick = {1},
                     xticklabels = {$\lambda_1$,$\lambda_2$},
                     yticklabels = {$1$}},
  softstyle/.style = {basestyle,
                      xtick = {\t},
                      ytick = {1},
                      xticklabels = {$\theta$},
                      yticklabels = {$1$}},
  garrotestyle/.style = {softstyle},
  hardstyle/.style = {softstyle}
}

\newcommand{\homdiff}
{%
  \begin{tikzpicture}
    \begin{axis}
    [
      homstyle,
      xlabel = {$\mArg$},
      ylabel = {$\mDiff(\mArg)$},
      ymin = -1.5,  
      ymax = 1.5, 
    ]
    
    \addplot[blue, very thick]
      plot (\x, {1});
      
    \end{axis}
  \end{tikzpicture}
}
  
\newcommand{\hompen}
{%
  \begin{tikzpicture}
    \begin{axis}
    [
      homstyle,
      xlabel = {$\mArg$},
      ylabel = {$\mPen(\mArg)$}
    ]
    
    \addplot[blue, very thick] 
      plot (\x, {\x^2});
      
    \end{axis}
  \end{tikzpicture}
}
  
\newcommand{\homshrink}
{%
  \begin{tikzpicture}
    \begin{axis}
    [
      homstyle,
      xlabel = {$\mArg$},
      ylabel = {$\mShrink(\mArg)$}
    ]
    
    \addplot[blue, very thick] 
      plot (\x, {0});
      
    \addplot[black, thin, dashed]
      plot (\x, \x);
      
    \end{axis}
  \end{tikzpicture}
}
  
\newcommand{\homact}
{%
  \begin{tikzpicture}
    \begin{axis}
    [
      homstyle,
      xlabel = {$\mArg$},
      ylabel = {$\mAct(\mArg)$}
    ]
    
    \addplot[blue, very thick] 
      plot (\x, {\x});
      
    \end{axis}
    
  \end{tikzpicture}
}

\newcommand{\charbdiff}[1]
{%
  \begin{tikzpicture}
    \def\la{#1}
    
    \begin{axis}
    [
      charbstyle,
      xlabel = {$\mArg$},
      ylabel = {$\mDiff(\mArg)$},
      ymin = -1.5,  
      ymax = 1.5, 
    ]
    
    \addplot[blue, very thick] 
      plot (\x, {1 / sqrt(1 + \x^2 / \la^2)});
    
    \end{axis}
  \end{tikzpicture}
}
    
\newcommand{\charbpen}[1]
{%
  \begin{tikzpicture}
    \def\la{#1}
    \pgfmathsetmacro{\ymarki}{2 * \la * \la * (sqrt(2) - 1)}
    \begin{axis}
    [
      charbstyle,
      xlabel = {$\mArg$},
      ylabel = {$\mPen(\mArg)$},
      ymajorticks=false
    ]
    
    \addplot[blue, very thick]
      plot (\x, {2 * \la^2 * sqrt(1 + \x^2 / \la^2) - 2 *  \la^2});
    
    \end{axis}
  \end{tikzpicture}
}
    
\newcommand{\charbshrink}[1]
{%
  \begin{tikzpicture}
    \def\la{#1}

    \pgfmathsetmacro{\ymarki}{(sqrt(3) - 1)/sqrt(3) * \la}
    \begin{axis}
    [
      charbstyle,
      xlabel = {$\mArg$},
      ylabel = {$\mShrink(\mArg)$},
      ymajorticks=false
    ]
    
    \addplot[blue, very thick] 
      plot (\x, {\x * (1 - sqrt(\la^2 / (\la^2 + 2 * \x^2)))});
      
    \addplot[black, thin, dashed] 
      plot (\x, {\x});
    
    \end{axis}
  \end{tikzpicture}
}
    
\newcommand{\charbact}[1]
{%
  \begin{tikzpicture}
    \def\la{#1}
  
    \pgfmathsetmacro{\ymarki}{\la / sqrt(2)}
    \begin{axis}
    [
      charbstyle,
      xlabel = {$\mArg$},
      ylabel = {$\mAct(\mArg)$},
      ymin = -1,  
      ymax = 1, 
      ytick = {\ymarki},
      yticklabel = {$\frac{\lambda}{\sqrt{2}}$}
    ]
    
    \addplot[blue, very thick]
      plot (\x, {\x / sqrt(1 + \x^2 / \la^2)});
    
    \end{axis}
    
  \end{tikzpicture}
} 

\newcommand{\pmiidiff}[1]
{%
  \begin{tikzpicture}
    \def\la{#1}
    
    \begin{axis}
    [
      pmiistyle,
      xlabel = {$\mArg$},
      ylabel = {$\mDiff(\mArg)$},
      ymin = -1.5,  
      ymax = 1.5, 
    ]
    
    \addplot[blue, very thick] 
      plot (\x, {exp(-\x^2 / (2*\la^2))});
    
    \end{axis}
  \end{tikzpicture}
}

\newcommand{\pmiipen}[1]
{%
  \begin{tikzpicture}
    \def\la{#1}
    
    \pgfmathsetmacro{\ymarki}{\la * \la * ln(2)}
    \begin{axis}
    [
      pmiistyle,
      xlabel = {$\mArg$},
      ylabel = {$\mPen(\mArg)$},
      ymax = 1, 
      ymin = -1, 
      ymajorticks=false
    ]
    
    \addplot[blue, very thick]
      plot (\x, {2 * \la^2 * (1 - exp(-\x^2 / (2*\la^2)))});
    
    \end{axis}
  \end{tikzpicture}
}

\newcommand{\pmiishrink}[1]
{%
  \begin{tikzpicture}
    \def\la{#1}
    \pgfmathsetmacro{\ymarki}{2/3 * \la}
    \begin{axis}
    [
      pmiistyle,
      xlabel = {$\mArg$},
      ylabel = {$\mShrink(\mArg)$},
      ymajorticks=false
    ]
    
    \addplot[blue, very thick] 
      plot (\x, {\x * (1 - exp(-\x^2 / (\la^2)))});
      
    \addplot[black, thin, dashed] 
      plot (\x, {\x});
    
    \end{axis}
  \end{tikzpicture}
}

\newcommand{\pmiiact}[1]
{%
  \begin{tikzpicture}
    \def\la{#1}
        
    \pgfmathsetmacro{\ymarki}{\la / sqrt(exp(1))}
    \begin{axis}
    [
      pmiistyle,
      xlabel = {$\mArg$},
      ylabel = {$\mAct(\mArg)$},
      ymax = 0.8, 
      ymin = -0.8, 
      ytick = {\ymarki},
      yticklabel = {$\frac{\lambda}{\sqrt{e}}$}
    ]
    
    \addplot[blue, very thick]
      plot (\x, {\x * exp(-\x^2 / (2*\la^2))});
    
    \end{axis}
    
  \end{tikzpicture}
}

\newcommand{\softdiff}[1]
{%
  \begin{tikzpicture}
    \def\t{#1}
  
    \pgfmathsetmacro{\xmarki}{sqrt(2) * \t}
    \pgfmathsetmacro{\xmarkii}{2 * \t^2}
    
    \begin{axis}
    [
      softstyle,
      xtick = {\xmarki},
      xticklabels = {$\sqrt{2}\,\theta$},
      xlabel = {$\mArg$},
      ylabel = {$\mDiff(\mArg)$},
      xmax = 3, 
      xmin = -3, 
      ymin = -1.5,  
      ymax = 1.5, 
      domain = -3:3
    ]
    
    \addplot[blue, very thick] 
      plot [domain=-10:-\t*sqrt(2)] (\x, {sqrt(2) * \t / abs(\x)});
    
    \addplot[blue, very thick] 
      plot [domain=\t*sqrt(2):10] (\x, {sqrt(2) * \t / abs(\x)});
    
    \addplot[blue, very thick] 
      plot [domain=-\t*sqrt(2):\t*sqrt(2)]   (\x, {1});
    
    \end{axis}
  \end{tikzpicture}
}

\newcommand{\softpen}[1]
{%
  \begin{tikzpicture}
    \def\t{#1}
    
    \pgfmathsetmacro{\xmarki}{sqrt(2) * \t}
    \pgfmathsetmacro{\xmarkii}{2 * \t^2}
    
    \begin{axis}
    [
      softstyle,   
      xtick = {\xmarki},
      xticklabels = {$\sqrt{2}\,\theta$},
      ytick = {\xmarkii},
      yticklabels = {$2\,\theta^2$},
      xlabel = {$\mArg$},
      ylabel = {$\mPen(\mArg)$},
      ymin = -1, 
      ymax = 1, 
      xmin = -1.4, 
      xmax = 1.4, 
    ]
    
    \addplot[blue, very thick] 
      plot [domain=-10:-\t*sqrt(2)] 
        (\x, {2 * sqrt(2) * \t * abs(\x) - 2 * \t^2});
    
    \addplot[blue, very thick] 
      plot [domain=\t*sqrt(2):10] 
        (\x, {2 * sqrt(2) * \t * abs(\x) - 2 * \t^2});
    
    \addplot[blue, very thick] 
      plot [domain=-\t*sqrt(2):\t*sqrt(2)]
        (\x, {\x^2});

    \end{axis}
  \end{tikzpicture}
}

\newcommand{\softshrink}[1]
{%
  \begin{tikzpicture}
    \def\t{#1}
    
    \pgfmathsetmacro{\xmarki}{sqrt(2) * \t}
    \pgfmathsetmacro{\xmarkii}{2 * \t^2}
    
    \begin{axis}
    [
      softstyle,    
      xlabel = {$\mArg$},
      ylabel = {$\mShrink(\mArg)$},
      ytick = \empty,
    ]
    
    \addplot[blue, very thick] 
      plot [domain=-10:-\t] (\x, {\x + \t});
    
    \addplot[blue, very thick] 
      plot [domain=\t:10] (\x, {\x - \t});
    
    \addplot[blue, very thick] 
      plot [domain=-\t:\t]   (\x, {0});
    
    \addplot[dashed, thin, black] 
      plot  (\x, \x);
    
    \end{axis}
  \end{tikzpicture}
}
 
\newcommand{\softact}[1]
{%
  \begin{tikzpicture}
    \def\t{#1}
   
    \pgfmathsetmacro{\xmarki}{sqrt(2) * \t}
    \pgfmathsetmacro{\xmarkii}{2 * \t^2}
   
    \begin{axis}
    [
      softstyle,   
      xtick = {\xmarki},
      xticklabels = {$\sqrt{2}\,\theta$},
      ytick = {\xmarki},
      yticklabels = {$\sqrt{2}\,\theta$},
      xlabel = {$\mArg$},
      ylabel = {$\mAct(\mArg)$}
    ]

    \addplot[blue, very thick] 
      plot [domain=-10:-\t*sqrt(2)] (\x, {sqrt(2) * \t * (-1)});
    
    \addplot[blue, very thick] 
      plot [domain=\t*sqrt(2):10] (\x, {sqrt(2) * \t *(1)});
    
    \addplot[blue, very thick] 
      plot [domain=-\t*sqrt(2):\t*sqrt(2)]   (\x, {\x});
    
    \end{axis}
    
  \end{tikzpicture}
}

\newcommand{\garrotediff}[1]
{%
  \begin{tikzpicture}
    \def\t{#1}
    
    \pgfmathsetmacro{\xmarki}{sqrt(2) * \t}
    \pgfmathsetmacro{\xmarkii}{2 * \t^2}
    
    \begin{axis}
    [
      garrotestyle, 
      xtick = {\xmarki},
      xticklabels = {$\sqrt{2}\,\theta$},
      xlabel = {$\mArg$},
      ylabel = {$\mDiff(\mArg)$},
      xmax = 3, 
      xmin = -3, 
      ymin = -1.5,  
      ymax = 1.5, 
      domain = -3:3
    ]
    
    \addplot[blue, very thick] 
      plot [domain=-10:-\t*sqrt(2)] (\x, {2 * \t^2 / (\x^2)});
    
    \addplot[blue, very thick] 
      plot [domain=\t*sqrt(2):10] (\x, {2 * \t^2 / (\x^2)});
    
    \addplot[blue, very thick] 
      plot [domain=-\t*sqrt(2):\t*sqrt(2)]   (\x, {1});
    
    \end{axis}
  \end{tikzpicture}
}

\newcommand{\garrotepen}[1]
{%
  \begin{tikzpicture}
    \def\t{#1}
    
    \pgfmathsetmacro{\xmarki}{sqrt(2) * \t}
    \pgfmathsetmacro{\xmarkii}{2 * \t^2}
    
    \begin{axis}
    [
      garrotestyle, 
      xtick = {\xmarki},
      xticklabels = {$\sqrt{2}\,\theta$},
      ytick = {\xmarkii},
      yticklabels = {$2\,\theta^2$},
      xlabel = {$\mArg$},
      ylabel = {$\mPen(\mArg)$},
      ymax = 3, 
      ymin = -3, 
      xmax = 2.5, 
      xmin = -2.5, 
      domain = -3:3
    ]
    
    \addplot[blue, very thick] 
      plot [domain=-10:-\t*sqrt(2)] 
        (\x, {2 * \t^2 * (ln(\x^2) - ln(2*\t^2) + 1)});
    
    \addplot[blue, very thick] 
      plot [domain=\t*sqrt(2):10] 
        (\x, {2 * \t^2 * (ln(\x^2) - ln(2*\t^2) + 1)});
    
    \addplot[blue, very thick] 
      plot [domain=-\t*sqrt(2):\t*sqrt(2)]  
        (\x, {\x^2});

    \end{axis}
  \end{tikzpicture}
}
    
\newcommand{\garroteshrink}[1]
{%
  \begin{tikzpicture}
    \def\t{#1}
    
    \pgfmathsetmacro{\xmarki}{sqrt(2) * \t}
    \pgfmathsetmacro{\xmarkii}{2 * \t^2}
    
    \begin{axis}
    [
      garrotestyle, 
      xlabel = {$\mArg$},
      ylabel = {$\mShrink(\mArg)$},
      ytick = \empty,
    ]
    
    \addplot[blue, very thick] 
      plot [domain=-10:-\t] (\x, {\x - \t^2/\x});
    
    \addplot[blue, very thick] 
      plot [domain=\t:10] (\x, {\x - \t^2/\x});
    
    \addplot[blue, very thick] 
      plot [domain=-\t:\t]   (\x, {0});
    
    \addplot[dashed, thin, black] 
      plot  (\x, \x);
    
    \end{axis}
  \end{tikzpicture}
}
    
\newcommand{\garroteact}[1]
{%
  \begin{tikzpicture}
    \def\t{#1}
    
    \pgfmathsetmacro{\xmarki}{sqrt(2) * \t}
    \pgfmathsetmacro{\xmarkii}{2 * \t^2}
    
    \begin{axis}
    [
      garrotestyle, 
      xtick = {\xmarki},
      xticklabels = {$\sqrt{2}\,\theta$},
      ytick = {\xmarki},
      yticklabels = {$\sqrt{2}\,\theta$},
      xlabel = {$\mArg$},
      ylabel = {$\mAct(\mArg)$}
    ]

    \addplot[blue, very thick] 
      plot [domain=-10:-\t*sqrt(2)] (\x, {2 * \t^2 / \x});
    
    \addplot[blue, very thick] 
      plot [domain=\t*sqrt(2):10] (\x, {2 * \t^2 / \x});
    
    \addplot[blue, very thick] 
      plot [domain=-\t*sqrt(2):\t*sqrt(2)]   (\x, {\x});
    
    \end{axis}
    
    \end{tikzpicture}
}

\newcommand{\harddiff}[1]
{%
  \begin{tikzpicture}
    \def\t{#1}
    
    \pgfmathsetmacro{\xmarki}{sqrt(2) * \t}
    \pgfmathsetmacro{\xmarkii}{2 * \t^2}
    
    \begin{axis}
    [
      hardstyle,          
      xtick = {\xmarki},
      xticklabels = {$\sqrt{2}\,\theta$},
      xlabel = {$\mArg$},
      ymin = -1.5,  
      ymax = 1.5, 
      ylabel = {$\mDiff(\mArg)$}
    ]
    
    \addplot[blue, very thick] 
      plot [domain=-10:-\t*sqrt(2)] (\x, {0});
    
    \addplot[blue, very thick] 
      plot [domain=\t*sqrt(2):10] (\x, {0});
    
    \addplot[blue, very thick] 
      plot [domain=-\t*sqrt(2):\t*sqrt(2)]   (\x, {1});
    
    \draw [dashed, thin, blue] (axis cs: {\t*sqrt(2)},1) -- 
                               (axis cs: {\t*sqrt(2)},0); 
                               
    \draw [dashed, thin, blue] (axis cs: {-\t*sqrt(2)},1) -- 
                               (axis cs: {-\t*sqrt(2)},0); 
    
    \end{axis}
  \end{tikzpicture}
}
    
\newcommand{\hardpen}[1]
{%
  \begin{tikzpicture}
    \def\t{#1}
    
    \pgfmathsetmacro{\xmarki}{sqrt(2) * \t}
    \pgfmathsetmacro{\xmarkii}{2 * \t^2}
    
    \begin{axis}
    [
      hardstyle, 
      xtick = {\xmarki},
      xticklabels = {$\sqrt{2}\,\theta$},
      ytick = {\xmarkii},
      yticklabels = {$2\,\theta^2$},
      xlabel = {$\mArg$},
      ylabel = {$\mPen(\mArg)$}
    ]
    
    \addplot[blue, very thick] 
      plot [domain=-10:-\t*sqrt(2)] (\x, {2 * \t^2});
    
    \addplot[blue, very thick] 
      plot [domain=\t*sqrt(2):10] (\x, {2 * \t^2});
    
    \addplot[blue, very thick] 
      plot [domain=-\t*sqrt(2):\t*sqrt(2)]   (\x, {\x^2});
    
    \end{axis}
  \end{tikzpicture}
}

\newcommand{\hardshrink}[1]
{%
  \begin{tikzpicture}
    \def\t{#1}
    
    \pgfmathsetmacro{\xmarki}{sqrt(2) * \t}
    \pgfmathsetmacro{\xmarkii}{2 * \t^2}
    
    \begin{axis}
    [
      hardstyle, 
      ytick = {\t},
      yticklabel = {$\theta$},
      xlabel = {$\mArg$},
      ylabel = {$\mShrink(\mArg)$}
    ]
    
    \addplot[blue, very thick] 
      plot [domain=-10:-\t] (\x, {\x});
    
    \addplot[blue, very thick] 
      plot [domain=\t:10] (\x, {\x});
    
    \addplot[blue, very thick] 
      plot [domain=-\t:\t]   (\x, {0});
    
    \addplot[dashed, thin, black] 
      plot  (\x, \x);
    
    \draw [dashed, thin, blue] (axis cs: {\t},{\t}) -- 
                               (axis cs: {\t},0); 
    
    \draw [dashed, thin, blue] (axis cs: {-\t},{-\t}) -- 
                               (axis cs: {-\t},0); 
      
    \end{axis}
  \end{tikzpicture}
}
    
\newcommand{\hardact}[1]
{%
  \begin{tikzpicture}
    \def\t{#1}
    
    \pgfmathsetmacro{\xmarki}{sqrt(2) * \t}
    \pgfmathsetmacro{\xmarkii}{2 * \t^2}
    
    \begin{axis}
    [
      hardstyle, 
      xtick = {\xmarki},
      xticklabels = {$\sqrt{2}\,\theta$},
      ytick = {\xmarki},
      yticklabels = {$\sqrt{2}\,\theta$},
      xlabel = {$\mArg$},
      ylabel = {$\mAct(\mArg)$}
    ]

    \addplot[blue, very thick] 
      plot [domain=-10:-\t*sqrt(2)] (\x, {0});
    
    \addplot[blue, very thick] 
      plot [domain=\t*sqrt(2):10] (\x, {0});
    
    \addplot[blue, very thick] 
      plot [domain=-\t*sqrt(2):\t*sqrt(2)]   (\x, {\x});
    
    \draw [dashed, thin, blue] (axis cs: {\t*sqrt(2)},{\t*sqrt(2)}) -- 
                               (axis cs: {\t*sqrt(2)},0); 
                               
    \draw [dashed, thin, blue] (axis cs: {-\t*sqrt(2)},{-\t*sqrt(2)}) -- 
                               (axis cs: {-\t*sqrt(2)},0); 
    
    \end{axis}
    
  \end{tikzpicture}
}

\tikzset
{
  block/.style = {shape=rectangle, rounded corners,
                  draw, anchor=center, minimum height = 2em,
                  align=center, minimum width = 4em,
                  inner sep=1ex},
  operator/.style = {block, circle, inner sep = .5ex, minimum width = 0em},
}

\newcommand{\stencilthreebyone}[3]
{
  \ensuremath
  {
    \renewcommand{\arraystretch}{1.1}
    \begin{array}{|c|c|c|} \hline
      #1 & #2 & #3\\ \hline
    \end{array}
    \renewcommand{\arraystretch}{1.0}
  }
}

\newcommand{\edge}[2]{\draw (#1) edge node (TMP) {} (#2);}

\newcommand{\layername}[4]{\path (#1) -- node (TMP) {} (#2);
                           \draw node [left of=TMP, xshift=#4] {#3};}

\newcommand{\edgelabel}[1]{\draw node [right of=TMP, anchor=west, 
                           xshift=-2.2em] {#1};}

\newcommand{\skipedge}[3] {\draw (#1) -- +(#3,0) |-
                           node (TMP) {}
                           node [near start,right] {Id} (#2);}

\newcommand{\dx}{\partial_x}
\newcommand{\mDiff}{g}
\newcommand{\mShrink}{S}
\newcommand{\mPen}{\Psi}
\newcommand{\mAct}{\Phi}
\newcommand{\mArg}{r}

\newtheorem{mytheorem}{Proposition}

\begin{document}

\twocolumn[
\icmltitle{Translating Diffusion, Wavelets, and      
           Regularisation into Residual Networks}



\icmlsetsymbol{equal}{*}

\begin{icmlauthorlist}
\icmlauthor{Tobias Alt}{mia}
\icmlauthor{Joachim Weickert}{mia}
\icmlauthor{Pascal Peter}{mia}
\end{icmlauthorlist}

\icmlaffiliation{mia}{Mathematical Image Analysis Group,
                      Faculty of Mathematics and Computer Science,
                      Campus E1.7, Saarland University, 66041 Saarbr\"ucken, 
                      Germany}

\icmlcorrespondingauthor{Tobias Alt}{alt@mia.uni-saarland.de}
\icmlcorrespondingauthor{Joachim Weickert}{weickert@mia.uni-saarland.de}
\icmlcorrespondingauthor{Pascal Peter}{peter@mia.uni-saarland.de}

\icmlkeywords{Partial Differential Equations,
              Nonlinear Diffusion,
              Wavelet Shrinkage,
              Variational Methods,
              Activation Functions,
              Residual Networks}

\vskip 0.3in]



\printAffiliationsAndNotice{}  

\begin{abstract}
Convolutional neural networks (CNNs) often perform well, but their 
stability is poorly understood. To address this problem, we consider 
the simple prototypical problem of signal denoising, where classical
approaches such as nonlinear diffusion, wavelet-based methods and 
regularisation offer provable stability guarantees. To transfer such 
guarantees to CNNs, we interpret numerical approximations of these
classical methods as a specific residual network (ResNet) architecture.
This leads to a dictionary which allows to translate diffusivities,
shrinkage functions, and regularisers into activation functions, and
enables a direct communication between the four research communities.
On the CNN side, it does not only inspire new families of nonmonotone
activation functions, but also introduces intrinsically stable architectures 
for an arbitrary number of layers.
\end{abstract}


\section{Introduction}

In view of the undeniable success of deep learning approaches in all
areas of data science  \cite{LBBH98,LBH15,Sch15a,GBC16}, there is a 
strong need to put them on a solid ground by establishing their 
mathematical foundations.

An {\em analytic} way towards this ambitious goal is to express 
successful CNN architectures in terms of well-founded mathematical 
concepts. However, deep learning offers a plethora of design 
possibilities, and modern architectures may involve hundreds of
layers and millions of parameters. Thus, this way of reducing a
highly complex system to a simple and transparent mathematical model 
is not only very burdensome, but also bears the danger to lose 
performance of critical CNN features along the way.

An alternative, {\em synthetic} way uses well-established models 
that offer deep mathematical insights to build simple components of 
neural architectures which inherit these qualities. 
While this constructive road seems less stony, it has been explored 
surprisingly little.

In this paper, we follow the road less taken and drive its simplicity 
to the extreme. Thus, at this point it is not our intention to design 
highly sophisticated architectures that produce state-of-the-art 
results in benchmarks. We do not even present any experiments at all. 

\subsection{Our Contribution}
  Our goal is to gain theoretical 
  insights that can be useful to suggest neural architectures 
  that are simpler, more compact, involve less parameters, and benefit 
  from provable stability guarantees.
  
  We establish a comprehensive framework that for the first time
  allows to translate diffusion methods, wavelet approaches, and variational
  techniques simultaneously into a specific CNN architecture. The reason 
  for choosing three denoising techniques lies in the intrinsic stability 
  of denoising: Noise is a perturbation of the input data, which is
  not supposed to change the denoised output substantially. To maximise 
  transparency and notational simplicity, we restrict ourselves to the 1D 
  setting and choose particularly simple representatives in each class: 
  Perona--Malik diffusion, Haar wavelet shrinkage, smooth first order 
  variational models, and a single ResNet block. We show that discrete 
  formulations of all three denoising approaches can be expressed as 
  a specific ResNet block. It inherits its stability directly from the
  three denoising algorithms. Thus, a ResNet consisting only of these 
  blocks is stable for any number of layers.

  Whereas typical CNNs learn convolution weights and fix the nonlinear 
  activation function to a simple design, we proceed in the opposite way:
  We fix the convolution kernels and study various nonlinear activation 
  functions that are inspired by the diffusivities, shrinkage functions,
  and variational regularisers. For researchers from the diffusion,
  wavelet or variational communities, this introduces a dictionary that 
  allows them to translate their methods directly into CNN architectures. 
  Deep learning researchers will find hitherto unexplored nonmonotone 
  activation functions and new motivations for existing ones.
   
  Our results question two architectural principles behind
  CNNs that are usually taken for granted. One of our findings is the
  fact that antisymmetric activation functions can occur naturally. More 
  importantly, we also show that nonmonotone activation functions
  do not contradict even the most restrictive notions of stability and
  well-posedness.
  

\subsection{Related Work}
  We see our paper in the tradition of a number of recent contributions
  to the mathematical foundations of deep learning. For a detailed survey 
  that covers results until 2017, we refer to \cite{VBGS17}. The following
  review focuses on works that are relevant for our paper and does not 
  discuss other interesting aspects such as expressiveness 
  \cite{PMRM17,RT18,GKNV19} and information theoretic interpretations 
  \cite{ST17} of CNNs.

  The seminal work of \cite{BM13} employs wavelet operations to come up 
  with scattering networks that perform well on classification problems. 
  This has been the starting point for a number of wavelet-inspired CNNs; 
  see e.g.~\cite{WB17,FTH18,WL18,RGPF20} and the 
  references therein. Usually they exploit the spectral information 
  or the multiscale nature of wavelets. Our work, however, focuses on
  iterated shift-invariant wavelet shrinkage on a single scale and
  utilises its connection to diffusion processes \cite{MWS03a}. 

  Gaining insights into CNNs is possible by studying their 
  energy landscapes \cite{NH17,SM17,COOS18,LXTS18,DVSH18}, their optimality 
  conditions \cite{HV17}, and by interpreting them as regularising 
  architectures \cite{KGC17,UVL18,DKMB20,KEKP20}. They can also be 
  connected to incremental proximal gradient methods 
  \cite{KKHP17,CP20,BGKR19,HHNP20}, 
  that are adequate for nonsmooth optimisation problems. 
  In the present paper we advocate an interpretation in terms of iterated 
  smooth energy minimisation problems. We do not require proximal steps and 
  can exploit direct connections to diffusion filters \cite{SchW98,RSW99a}.

  A prominent avenue to establish mathematical foundations of CNNs is
  through their analysis in terms of stability. This can be achieved by 
  studying their invertibility properties \cite{BDFM18,CMHR18},
  by exploiting sparse coding concepts \cite{RASE20}, and by interpreting 
  deep learning as a parameter identification or optimal control problem 
  for ordinary differential equations \cite{HR17,TG19,ZS20}. CNNs can also be 
  connected to flows of diffeomorphisms \cite{RDF20} and to parabolic or 
  hyperbolic PDEs \cite{We17,LZLD17,LS18,SPBD20}, where it is possible to 
  transfer $L^2$ stability results \cite{RH20}. Our work focuses on diffusion 
  PDEs. They allow us to establish stricter stability notions such 
  as $L^\infty$ stability and sign stability. 
  
  Within practical settings, connections between PDEs
  and CNNs are the basis of trainable diffusion models for
  inverse problems \cite{CP16,AH20} and approaches for directly learning PDEs 
  in a data-driven manner \cite{Sch17,LLD19}. We on the other hand 
  specify the PDE directly and analyse its implications on the CNN side. 
  Furthermore, trainable variants of the nonlinearities that we investigate 
  have shown success for traditional models from the fields of diffusion 
  filtering \cite{BL20}, wavelet shrinkage \cite{HS08,SS14,AW20}, and 
  regularisation \cite{KP13, DSV17}.
    
  We argue for shifting the focus of CNN models towards more sophisticated 
  and also nonmonotone activation functions. The CNN literature offers 
  only few examples of training activation functions \cite{KSM18}
  or designing them in a well-founded and flexible way \cite{Un19}.
  Nonmonotone activation functions have been suggested already before
  the advent of deep learning \cite{FMNP93,MR94}, but fell into oblivion
  afterwards. We revitalise this idea by providing a natural justification
  from the theory of diffusion filtering.


\subsection{Organisation of the Paper}
Our paper is structured as follows. We review general formulations of 
nonlinear diffusion, wavelet shrinkage, variational regularisation,
and residual networks in Section \ref{sec_model_intro}.
In Section \ref{sec_connections}, we discuss numerical approximations for 
three instances of the classical models. We interpret them in terms of 
a specific residual network architecture, for which we derive explicit 
stability guarantees. This leads to a dictionary for translating the
nonlinearities of the three classical methods to activation functions,
which is presented in Section \ref{sec_discussion}. For a selection of 
the most popular nonlinearities, we derive their counterparts and discuss 
novel consequences for the design of neural networks in detail.
Finally, we summarise our conclusions in Section \ref{sec_conclusions}.


\section{Basic Approaches}\label{sec_model_intro}
This section sketches nonlinear diffusion, wavelet shrinkage, 
variational regularisation, and residual networks in a general fashion. 
For each model, we highlight and discuss its central design choice.

  To ensure a consistent notation, all models in this section produce an 
  output signal $u$ from an input signal $f$. We define continuous 
  one-dimensional signals $u$, $f$ as mappings from a signal domain 
  $\Omega = [a,b]$ to a codomain $[c,d]$. 
  We employ reflecting boundary conditions 
  on the signal domain boundaries $a$ and $b$. The discrete signals 
  $\bm u$, $\bm f \in \mathbb{R}^N$ are obtained by sampling the continuous 
  functions at $N$ equidistant positions with grid size $h$.
  
 
  \subsection{Nonlinear Diffusion}
    In nonlinear diffusion \cite{PM90}, filtered versions $u(x,t)$ of 
    an initial signal $f(x)$ are computed as solutions of the nonlinear 
    diffusion equation
    \begin{equation}\label{eq_nonlinear_diffusion}
      \partial_t u = \dx \left(
                           \mDiff\!\left(\dx u \right)  \dx u 
                         \right)
    \end{equation}
    with initial condition $u(x,0)=f(x)$ and diffusion time $t$. The evolution 
    creates gradually simplified versions of $f$. The central design choice 
    lies in the nonnegative, nonincreasing, and bounded 
    diffusivity $\mDiff(\mArg)$ which 
    controls the amount of smoothing depending on the local structure of 
    the evolving signal. Choosing 
    the constant diffusivity $\mDiff(\mArg) = 1$ \cite{Ii62} leads to a 
    homogeneous diffusion process that smoothes the signal equally at all 
    locations. A more sophisticated diffusivity such as the exponential 
    Perona--Malik diffusivity 
    $\mDiff(\mArg) = \exp \left(-\frac{r^2} {2\lambda^2} \right)$
    \cite{PM90} inhibits smoothing around discontinuities where $|\dx u|$ is 
    larger than the contrast parameter $\lambda$. This allows 
    discontinuity-preserving smoothing.
  
  \subsection{Wavelet Shrinkage}
    Classical discrete wavelet shrinkage \cite{DJ94} manipulates a discrete 
    signal $\bm f$ within a wavelet basis. With the following three-step 
    framework, one obtains a filtered signal $\bm u$:
    \begin{enumerate} 
        \item \textit{Analysis:} One transforms the input signal $\bm f$ to
              wavelet and scaling coefficients by a transformation $\bm W$.
        
        \item \textit{Shrinkage:} A scalar-valued shrinkage function
              $\mShrink(\mArg)$ with a threshold parameter $\theta$ is applied 
              component-wise to the wavelet coefficients. The scaling 
              coefficients remain unchanged.
        
        \item \textit{Synthesis:} One applies a back-transformation 
              $\bm{\tilde{W}}$ to the manipulated coefficients to obtain the 
              result $\bm u$:
        \begin{equation}
          \bm u = \bm {\tilde{W}} \mShrink\! \left(\bm W \bm f\right).
        \end{equation}
    \end{enumerate} 
    Besides the choice of the wavelet basis, the result is strongly influenced 
    by the shrinkage function $\mShrink(\mArg)$. The hard shrinkage function 
    \cite{Ma98} eliminates all coefficients with a magnitude smaller than the 
    threshold parameter, while the soft shrinkage function \cite{Do95} 
    additionally modifies the remaining coefficients equally.
  
    The classical wavelet transformation is not shift-invariant: 
    Transforming a 
    shifted input signal changes the resulting set of coefficients. 
    To this end, cycle spinning was proposed in \cite{CD95}, which 
    averages the results of wavelet shrinkage for all possible shifts of the 
    input signal. This shift-invariant wavelet transformation will serve 
    as a basis for connecting wavelet shrinkage to nonlinear diffusion.


  \subsection{Variational Regularisation}
    Variational regularisation \cite{Wh23, Ti63} pursues the goal of
    finding a function $u(x)$ that minimises an energy functional. A general 
    formulation of such a functional is given by
    \begin{equation}
      E(u) = \int_{\Omega} \Big( D(u,f) + \alpha  R(u) \Big)  \, dx,
    \end{equation}
    where a data term $D(u,f)$ drives $u$ towards the input data $f$, and a 
    regularisation term $R(u)$ enforces smoothness conditions on $u$. We can 
    control the balance between both terms by the regularisation parameter 
    $\alpha > 0$. A solution minimising the energy is found via the 
    Euler--Lagrange equations \cite{GF00a}. These are PDEs that describe 
    necessary conditions for a minimiser.

    A simple yet effective choice for the regularisation term is a first order 
    regularisation of the form $R(u) = \mPen\left(\dx u\right)$
    with a regulariser $\mPen(\mArg)$. The nonlinear function $\mPen$ 
    penalises variations in $\dx u$ to enforce a certain smoothness condition 
    on $u$. Choosing e.g. $\mPen(\mArg) = \mArg^2$ is called 
    Whittaker--Tikhonov regularisation \cite{Wh23, Ti63}.


  \subsection{Residual Networks} 
    Residual networks \cite{HZRS16} are a popular CNN architecture as they are 
    easy to train, even for a high number of network layers. They consist of 
    chained residual blocks. A residual block is made up of two convolutional 
    layers with biases and nonlinear activation functions after each 
    layer. Each block computes the output signal $\bm u$ from an input signal 
    $\bm f$ by
    \begin{equation}\label{eq_residual_block}
      \bm u = \sigma_2 \! \left(
                          \bm f + \bm W_2 \, \sigma_1 \! 
                                  \left(\bm W_1 \bm f + \bm b_1\right) 
                                + \bm b_2
                       \right),
    \end{equation}
    with discrete convolution matrices $\bm W_1, \bm W_2$, activation 
    functions $\sigma_1, \sigma_2$ and bias vectors $\bm b_1, \bm b_2$.
    
    The main difference to general feed-forward CNNs lies in the 
    skip-connection which adds the original input signal $\bm f$ to the result 
    of the inner activation function. This helps with the vanishing gradient 
    problem and substantially improves training performance.

    The crucial difference between residual networks and the three previous 
    approaches is the design focus: The three classical methods consider 
    complex nonlinear modelling functions, while CNNs mainly focus on 
    learning convolution weights and use simple activation functions. We will 
    see that by permitting more general activation functions, we can relate 
    all four methods within a unifying framework. 


\section{Translation into Residual Networks} \label{sec_connections}
  Now we are in a position to discuss numerical approximations for the 
  three classical models that allow to interpret them in terms of a 
  residual network architecture. 


  \subsection{From Nonlinear Diffusion to Residual Networks}
    \label{sec_diffusion_to_resnets}
    In practice, the continuous diffusion process is discretised and iterated 
    to approximate the continuous solution $u(x,T)$ for a stopping time $T$. 
    With the help of the flux function $\mAct(\mArg) = \mDiff(\mArg) \, \mArg$
    we rewrite the diffusion equation \eqref{eq_nonlinear_diffusion} as
    \begin{equation}
      \partial_t u = \dx \left(\mAct \! \left(\dx u \right)\right).
    \end{equation}
    For this equation, we perform a standard discretisation in the spatial and 
    the temporal domain. This yields an explicit scheme which can be iterated. 
    Starting with an initial signal $\bm u^0=\bm f=(f_i)_{i=1}^N$, the 
    evolving signal $\bm{u}^k$ at a time step $k$ is used to compute 
    $\bm u^{k+1}$ at the next step by
    \begin{equation}\label{eq_nonlinear_diffusion_discrete}
    \resizebox{0.90\linewidth}{!}{$
      \frac{u^{k+1}_i - u^k_i}{\tau} = \frac{1}{h}
                   \left(\mAct\!\left(\frac{u^k_{i+1} - u^k_i}{h}\right) 
                         - \mAct\!\left(\frac{u^k_i - u^k_{i-1}}{h}\right)
                   \!\right).$
    }
    \end{equation}
    Here the temporal derivative is discretised by a forward difference 
    with time step size $\tau$. We apply a forward difference to
    implement the inner spatial derivative operator and a backward
    difference for the outer spatial derivative operator. Both can be
    realised with a simple convolution.


    \afterpage{
      \begin{table*}
        \caption{Dictionary for diffusivities $\mDiff(\mArg)$, regularisers 
                 $\mPen(\mArg)$, wavelet shrinkage functions $\mShrink(\mArg)$, 
                 and activation functions $\mAct(\mArg)$. A nonlinearity from a 
                 row can be translated into a nonlinearity from a column with 
                 the respective equation.  
                 \label{table_equivalences}}
        \vspace{2pt}
        
\setlength{\tabcolsep}{2pt}
\centering
\everymath{\displaystyle}
\resizebox{\textwidth}{!}{
\begin{tabular}{|c|c|c|c|c|}
    \hline
      \backslashbox{from}{to}
    & Diffusivity
    & Regulariser
    & Shrinkage Function
    & Activation Function
  \\ \hline  &&&& \\[-2ex]
      Diffusivity
    & $\mDiff(\mArg)$
    & $\mPen(\mArg) = 2
       \int\limits_{0}^{\mArg} \mDiff(x) \, x \, dx$
    & $\mShrink(\mArg) = 
       \mArg \left(
               1 - 4 \tau \, \mDiff \! \left(\sqrt{2} \mArg\right)
             \right)$
    & $\mAct(\mArg) = 
       \mDiff(\mArg)  \, \mArg$
  \\[3ex] \hline  &&&& \\[-2ex]
      Regulariser
    & $\mDiff(\mArg) = 
       \frac{\mPen^\prime(\mArg)}{2\mArg}$
    & $\mPen(\mArg)$
    & $\mShrink(\mArg) = 
       \mArg - \sqrt{2} \alpha \, \mPen^\prime \! \left(\sqrt{2} \mArg\right)$
    & $\mAct(\mArg) = 
       \frac12 \mPen^\prime(\mArg)$
  \\[1.5ex] \hline  &&&& \\[-2ex]
      \makecell{Shrinkage\\Function}
    & $\mDiff(\mArg) \!=\! 
       \frac{1}{4 \tau}\!\left(\! 
                         1 \!-\! \frac{\sqrt{2}}{\mArg} 
                             \mShrink\!\left( 
                                       \frac{\mArg}{\sqrt{2}}
                                     \right)\!\!
                       \right)\!$
    & $\mPen(\mArg) \!=\! 
       \frac{1}{4 \alpha} 
         \!\left(\! 
           \mArg^2 \!-\! 2\sqrt{2}
                         \int\limits_{0}^{\mArg}\!
                           \mShrink\!\left(\frac{x}{\sqrt{2}}\right) \! dx
           \right)\!$
    & $\mShrink(\mArg)$
    & $\mAct(\mArg) \!=\! 
       \frac{1}{4 \tau} \!\left(\! 
                          \mArg \!-\! \sqrt{2} 
                                  \,\mShrink\! \left( 
                                             \frac{\mArg}{\sqrt{2}}
                                           \right)\!\!
                        \right)\!$
  \\[3ex] \hline  &&&& \\[-2ex]
      \makecell{Activation\\Function}
    & $\mDiff(\mArg) = 
       \frac{\mAct(\mArg)}{\mArg}$
    & $\mPen(\mArg) = 2
       \int\limits_{0}^{\mArg} \mAct(x) \, dx$ 
    & $\mShrink(\mArg) =
               \mArg \!-\! 2 \sqrt{2} \tau \, \mAct\left(\sqrt{2} \mArg\right)$
    & $\mAct(\mArg)$
  \\[3ex] \hline
\end{tabular}
}

      \end{table*}
    }
    

    To obtain a scheme which is stable in the $L^\infty$ norm, one can
    show that the time step size must fulfil
    \begin{equation} \label{eq:stab}
     \tau \leq \frac{h^2}{2 \mDiff_\text{max}},
    \end{equation}
    where $\mDiff_\text{max}$ is the maximum value that the diffusivity
    $\mDiff(\mArg) = \frac{\mAct(\mArg)}{\mArg}$ can attain \cite{We97}. 
    This guarantees a maximum--minimum principle, stating that
    the values of the filtered signal $\bm u^k$ do not lie
    outside the range of the original signal~$\bm f$.

    To achieve a substantial filter effect, one often needs
    a diffusion time $T$ that exceeds the stability limit in (\ref{eq:stab}).
    Then one concatenates $m$ explicit steps with a time step size 
    $\tau=\frac{T}{m}$ that satisfies (\ref{eq:stab}).

    In order to translate diffusion into residual networks, we rewrite 
    the explicit scheme \eqref{eq_nonlinear_diffusion_discrete} in 
    matrix-vector form:
    \begin{equation}\label{eq_residual_block_diffusion}
      \bm u^{k+1} = \bm u^k + \tau \bm D^-_h 
                      \left(
                        \mAct \! \left(\bm D^+_h \bm u^k\right)
                      \right),
    \end{equation}
    where $\bm D^+_h$ and $\bm D^-_h$ are convolution matrices denoting 
    forward and backward difference operators with grid size $h$, 
    respectively. In this notation, the resemblance to a residual block 
    becomes apparent:
    \begin{mytheorem}\label{theo_stab}
    A diffusion step \eqref{eq_residual_block_diffusion} is equivalent 
    to a residual block \eqref{eq_residual_block} if
    \begin{equation}
      \sigma_1 = \mAct, \quad
      \sigma_2 = \textup{Id}, \quad
      \bm W_1 = \bm D^+_h, \quad
      \bm W_2 = \tau  \bm D^-_h,
    \end{equation}
    and the bias vectors $\bm{b}_1$, $\bm{b}_2$ are set to $\bm 0$. 
    \end{mytheorem}
    We see that the convolutions implement forward and backward difference 
    operators. Crucially, the inner activation function $\sigma_1$ corresponds 
    to the flux function $\mAct$. The effect of the skip-connection in the 
    residual block also becomes clear now: It is the central ingredient to 
    realise a time discretisation. We call a block of this form a 
    \emph{diffusion block}.
 

    \begin{figure}
      \begin{center}
        \begin{tikzpicture}[-latex]
  \matrix (m)
  [
    matrix of nodes,
    column 1/.style = {nodes={block}}
  ]
  {
        |(input)|          
        $u^k_i$
    \\[5.5ex]              
        |(fwd)|
        $\frac{u^k_{i+1} - u^k_i}{h}$%
    \\[5.5ex]
        |(flux)|
        $\mAct\left(\frac{u^k_{i+1} - u^k_i}{h}\right)$%
    \\[5.5ex]
        |(div)|
        $\frac{\tau}{h}
           \left(  \mAct\left(\frac{u^k_{i+1} - u^k_i}{h}\right) 
                 - \mAct\left(\frac{u^k_i - u^k_{i-1}}{h}\right)\right)$%
    \\[2ex]
        |[operator] (add)|
        $+$%
    \\[2ex]
        |(output)|         
        $u^{k+1}_i$%
    \\
  };

    \def\nameshift{-3em}                    
    \def\skipconnshift{3.5}                   
  
    \layername{input}{fwd}{Convolution}{\nameshift}   
  
    \edge{input}{fwd}                                
    \edgelabel{$\ast\, \frac{1}{h} \cdot \stencilthreebyone{0}{-1}{1}$}  
  
    \layername{fwd}{flux}{Activation}{\nameshift}      
  
    \edge{fwd}{flux}                          
    \edgelabel{$\mAct(\mArg) = \mDiff(\mArg) \, \mArg$}
                                                
    \layername{flux}{div}{Convolution}{\nameshift}      
  
    \edge{flux}{div}                          
    \edgelabel{$\ast\, \frac{\tau}{h} \cdot \stencilthreebyone{-1}{1}{0}$}  
  
    \edge{div}{add}                             
  
    \edge{add}{output}
  
    \skipedge{input}{add}{\skipconnshift}   

    \node[above of=input, yshift=0] (TMP) {};
    \draw (TMP) -- (input);

    \node[below of=output, yshift=0] (TMP) {};
    \draw (output) -- (TMP);

    \node[left of=input, xshift=\nameshift-3em, yshift=2em] (topleft) {};
    \node[right of=output, xshift=9em, yshift=-1.6em] (bottomright) {};
    \draw (topleft) rectangle (bottomright);

    \node [left of=flux, rotate=90, anchor=north, yshift=8em]
          {\textbf{Diffusion Block}};
\end{tikzpicture}
      \end{center}
      \vspace{-2em}
      
      \caption{Diffusion block for one explicit nonlinear
               diffusion step \eqref{eq_nonlinear_diffusion_discrete} 
               with flux or activation function $\mAct(\mArg)$ and 
               time step size $\tau$. Stencils denote discrete 
               convolution weights centred in position $i$.
               \label{fig_diffusion_block}}
    \end{figure}
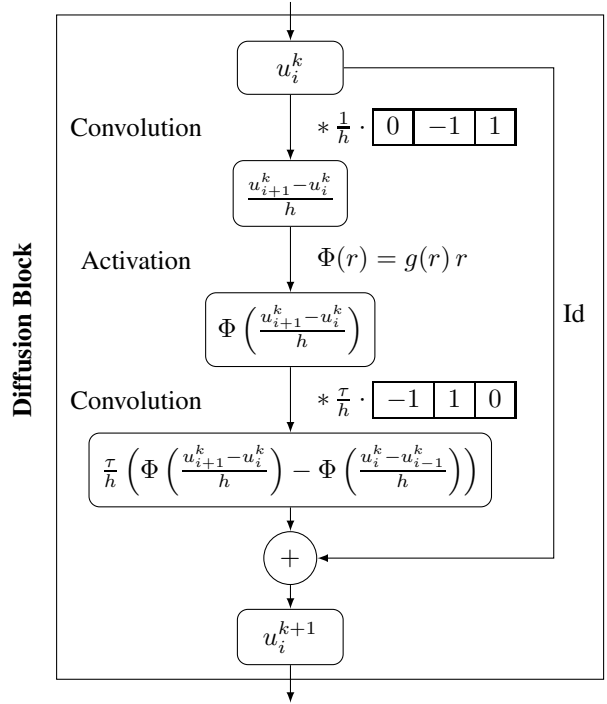

    Figure \ref{fig_diffusion_block} visualizes such a diffusion block. Graph 
    nodes contain the current state of the signal at position $i$, while edges 
    describe operations which are applied to proceed from one node to the next.

    A related approach for learning parameters of nonlinear 
    diffusion filters in the context of inverse problems has been proposed 
    in \cite{CYP15}. However, their 
    work establishes a diffusion-reaction framework: The skip-connection 
    rewards similarity to the original image in each step. Therefore, it cannot 
    be translated directly into a residual block \eqref{eq_residual_block}.
    On the contrary, we use a pure diffusion model. Our skip-connection rewards 
    similarity to the image from the previous layer and is thus compatible with 
    the residual block formulation.
    
    A consequence of Proposition \ref{theo_stab} is that a 
    residual network chaining $m$ diffusion blocks with activation function 
    $\mAct(\mArg)$ and time step size $\tau$ approximates a nonlinear diffusion 
    process with stopping time $T=m\tau$ and diffusivity 
    $\mDiff(\mArg)=\frac{\mAct(\mArg)}{\mArg}$. These insights enable us to 
    translate a diffusivity $\mDiff$ directly into an activation function 
    that coincides with the flux function $\Phi$:
    \begin{equation}\label{eq_diff_act_connection}
    \setlength{\fboxsep}{2mm}
    \setlength{\fboxrule}{0.5mm}
    \fbox{
       $\mAct(\mArg) \;=\; \mDiff(\mArg) \: \mArg$
     }
    \end{equation}
    While this translation from a diffusion step to a residual block 
    appears simple, it will serve as the Rosetta stone in our dictionary:
    It also allows to connect wavelet methods and variational regularisation  
    to residual networks, since both paradigms can be related to      
    diffusion \cite{MWS03a,SchW98}. To keep our paper self-contained,
    let us now sketch these correspondences.


  \subsection{From Wavelet Shrinkage to Nonlinear Diffusion}
    \label{sec_shrinkage_stability}
    The connection between diffusion PDEs and specific ResNets can be extended 
    to wavelet shrinkage. This is possible due to various equivalence results 
    between diffusion methods and wavelets; see 
    e.g.~\cite{MWS03a,WSMW05,MP07,WSW07,DJS17}. In our context, it is 
    sufficient 
    to focus on the results of \cite{MWS03a}.
    
    To explore the connection between wavelet shrinkage and nonlinear 
    diffusion, \cite{MWS03a} consider shift-invariant Haar wavelet 
    shrinkage on the finest scale. The missing multiscale structure
    is compensated by iterating this shrinkage. They show that one 
    step with shrinkage function $\mShrink(\mArg)$ is equivalent
    to an explicit diffusion step with diffusivity $\mDiff(\mArg)$,
    grid size $h=1$, and time step size $\tau$ if
    \begin{align}\label{eq_diffusion_wavelet}
      \mDiff(\mArg) = \frac{1}{4\tau}
                      \left(
                        1 - \frac{\sqrt{2}}{\mArg}
                            \mShrink \! \left(\frac{\mArg}{\sqrt{2}}\right)
                      \right).
    \end{align}
    The $L^\infty$ stability condition (\ref{eq:stab}) from the diffusion
    case translates into a condition on the shrinkage function
    \begin{equation}\label{eq_shrinkage_maxmin_stability}
      -\mArg \leq S(\mArg) \leq \mArg \quad \text{for } \mArg > 0,
    \end{equation}
    which is less restrictive than the typical design principle 
    \begin{equation}\label{eq_shrinkage_sign_stability}
      \phantom{-} 0 \leq S(\mArg) \leq \mArg \quad \text{for } \mArg > 0.
    \end{equation}
    Mr\'azek et al. show that the latter one leads to a sign stable
    process in the sense of \cite{Sch30}, i.e. the resulting signal
    shows not more sign changes
    than the input signal. This is a stronger stability notion than
    $L^\infty$ stability. It limits the time step size to
    $\tau \leq \frac{h^2}{4 \mDiff_\text{max}}$. This is half the 
    bound of (\ref{eq:stab}).

    
    \afterpage{
      \begin{table*}
        \caption{Function plots for selected diffusivities, regularisers,       
                 shrinkage functions, and activation functions. 
                 The names of known functions are written above the graphs. 
                 Bold font indicates the best known function for each row. Axes 
                 and parameters are individually scaled for optimal qualitative 
                 inspection.
                 \label{table_plots}} 
        \vspace{8pt}
        \centering
\def\plotwidth{0.2\linewidth}
\renewcommand{\arraystretch}{1.3}
\begin{tabular}{|c|c|c|c|}
  \hline 
      Diffusivity $\mDiff(\mArg)$
    & Regulariser $\mPen(\mArg)$
    & Shrinkage Function $\mShrink(\mArg)$
    & Activation Function $\mAct(\mArg)$
  \\ \hline\multicolumn{4}{c}{}\\[-0.7em]\hline
      \textbf{Constant}
    & Whittaker--Tikhonov
    & 
    & Identity
  \\
      \resizebox{\plotwidth}{!}{\homdiff}
    & \resizebox{\plotwidth}{!}{\hompen}
    & \resizebox{\plotwidth}{!}{\homshrink}
    & \resizebox{\plotwidth}{!}{\homact}
  \\ \hline
      \textbf{Charbonnier}
    & 
    &
    & 
  \\ 
      \resizebox{\plotwidth}{!}{\charbdiff{0.6}}
    & \resizebox{\plotwidth}{!}{\charbpen{0.6}}        
    & \resizebox{\plotwidth}{!}{\charbshrink{1.2}}     
    & \resizebox{\plotwidth}{!}{\charbact{0.6}}
  \\ \hline
      Truncated TV
    & Huber
    & \textbf{Soft}
    & 
  \\
      \resizebox{\plotwidth}{!}{\softdiff{0.75}}
    & \resizebox{\plotwidth}{!}{\softpen{0.4}}        
    & \resizebox{\plotwidth}{!}{\softshrink{0.75}}
    & \resizebox{\plotwidth}{!}{\softact{0.75}}
  \\ \hline\multicolumn{4}{c}{}\\[-0.7em]\hline
      \textbf{Perona--Malik}
    & 
    & 
    & 
  \\ 
      \resizebox{\plotwidth}{!}{\pmiidiff{0.6}}
    & \resizebox{\plotwidth}{!}{\pmiipen{0.6}}
    & \resizebox{\plotwidth}{!}{\pmiishrink{1.2}}     
    & \resizebox{\plotwidth}{!}{\pmiiact{0.6}}
  \\ \hline 
      Truncated BFB
    & 
    & \textbf{Garrote}
    &
  \\
      \resizebox{\plotwidth}{!}{\garrotediff{0.75}}
    & \resizebox{\plotwidth}{!}{\garrotepen{0.75}}
    & \resizebox{\plotwidth}{!}{\garroteshrink{0.75}}
    & \resizebox{\plotwidth}{!}{\garroteact{0.75}}
  \\ \hline 

    & Truncated Quadratic
    & \textbf{Hard}
    &
  \\
      \resizebox{\plotwidth}{!}{\harddiff{0.75}}
    & \resizebox{\plotwidth}{!}{\hardpen{0.75}}
    & \resizebox{\plotwidth}{!}{\hardshrink{0.75}}
    & \resizebox{\plotwidth}{!}{\hardact{0.75}}
  \\ \hline 
\end{tabular}
      \end{table*}
  
      \begin{landscape}
        \begin{table}
          \caption{Formulas for the 
                   function plots in Table \ref{table_plots}. The names of 
                   known functions are written above the equations. Bold font 
                   indicates the best known function for each row.
                   \label{table_equations}}
          \vspace{8pt}
          \centering
\setlength{\tabcolsep}{4pt}
\everymath{\displaystyle}
\renewcommand{\arraystretch}{1.3}
\begin{tabular}{|c|c|c|c|}
  \hline 
      Diffusivity $\mDiff(\mArg)$
    & Regulariser $\mPen(\mArg)$
    & Shrinkage Function $\mShrink(\mArg)$
    & Activation Function $\mAct(\mArg)$
  \\ \hline\multicolumn{4}{c}{}\\[-0.7em]\hline
      \textbf{Constant}
    & Whittaker--Tikhonov
    & 
    & Identity
  \\
      $\mDiff(\mArg) = 1$
    & $\mPen(\mArg) = \mArg^2$
    & $\mShrink(\mArg) = 0$
    & $\mAct(\mArg) = \mArg$
  \\ \hline
      \textbf{Charbonnier}
    & 
    &
    & 
  \\ 
      $\mDiff(\mArg) = \frac{1}{\sqrt{1 + \frac{\mArg^2}{\lambda^2}}}$
    & $\mPen(\mArg) =   2 \lambda^2 \sqrt{1 + \frac{\mArg^2}{\lambda^2}}        
                      - 2 \lambda^2$
    & $\mShrink(\mArg) 
        = \mArg \left( 
                  1 - \frac{1}{\sqrt{1 + \frac{2\mArg^2}{\lambda^2}}}
                \right)$
    & $\mAct(\mArg) = \frac{\mArg}{\sqrt{1 + \frac{\mArg^2}{\lambda^2}}}$
  \\[4ex] \hline
    Truncated TV
    & Huber
    & \textbf{Soft}
    & 
  \\
      $\mDiff(\mArg)
        = \left\{\!
            \begin{array}{ll}
              1                         
                ,&  \!|\mArg| \leq \sqrt{2}\,\theta  \\
              \frac{\sqrt{2}\,\theta}{|\mArg|}  
                ,&  \!|\mArg| >    \sqrt{2}\,\theta
            \end{array}
          \right.$     
    & $\mPen(\mArg) 
        = \left\{\!
            \begin{array}{ll}
              \mArg^2                 
                ,&  \!|\mArg| \leq \sqrt{2}\,\theta  \\
              2 \theta \left(\sqrt{2} \, |\mArg| - \theta\right) 
                ,&  \!|\mArg| >    \sqrt{2}\,\theta
            \end{array} 
          \right.$
    & $\mShrink(\mArg) 
        = \left\{\!
            \begin{array}{ll}
              0                       
                ,&  \!|\mArg| \leq \theta            \\
              \mArg - \theta \, \text{sgn}(\mArg)  
                ,&  \!|\mArg| >    \theta          
            \end{array}
          \right.$
    & $\mAct(\mArg) 
        = \left\{\!
            \begin{array}{ll}
              \mArg                  
                ,&  \!|\mArg| \leq \sqrt{2}\,\theta  \\
              \sqrt{2}\,\theta\, \text{sgn}(\mArg)
                ,&  \!|\mArg| >    \sqrt{2}\,\theta
            \end{array}
          \right.$
  \\[4ex] \hline\multicolumn{4}{c}{}\\[-0.7em]\hline
      \textbf{Perona--Malik}
    & 
    & 
    & 
  \\ 
      $\mDiff(\mArg) = \exp \! \left(-\frac{\mArg^2}{2\lambda^2}\right)$
    & $\mPen(\mArg) = 2 \lambda^2
                      \left(
                        1 - \exp \! \left(-\frac{\mArg^2}{2\lambda^2}\right)
                      \right)  $
    & $\mShrink(\mArg) = \mArg \left(
                                 1 -\exp \! 
                                 \left(-\frac{\mArg^2}{\lambda^2}\right)
                               \right)$
    & $\mAct(\mArg) = \mArg \exp \! \left(-\frac{\mArg^2}{2\lambda^2}\right)$
  \\[2ex] \hline
      Truncated BFB
    & 
    & \textbf{Garrote}
    &
  \\
      $\mDiff(\mArg)
        = \left\{\!
            \begin{array}{ll}
              1                         
                ,&  \!|\mArg| \leq \sqrt{2}\,\theta  \\
              \frac{2\theta^2}{\mArg^2}  
                ,&  \!|\mArg| >    \sqrt{2}\,\theta
            \end{array}
          \right.$     
    & $\mPen(\mArg) 
        = \left\{\!
            \begin{array}{ll}
              \mArg^2  
                  ,& \!|\mArg| \leq \sqrt{2}\,\theta  \\
                 2  \theta^2 
                  \left(
                    \ln \! \left(\frac{\mArg^2}{2\theta^2}\right) + 1
                  \right)
                  ,& \!|\mArg| >    \sqrt{2}\,\theta  \\
            \end{array} 
          \right.$
    & $\mShrink(\mArg) 
        = \left\{\!
            \begin{array}{ll}
              0                       
                ,&  \!|\mArg| \leq \theta            \\
              \mArg - \frac{\theta^2}{\mArg}  
                ,&  \!|\mArg| >    \theta          
            \end{array}
          \right.$
    & $\mAct(\mArg) 
        = \left\{\!
            \begin{array}{ll}
              \mArg                  
                ,&  \!|\mArg| \leq \sqrt{2}\,\theta  \\
              \frac{2\theta^2}{\mArg}  
                ,&  \!|\mArg| >    \sqrt{2}\,\theta
            \end{array}
          \right.$
  \\[4ex] \hline

    & Truncated Quadratic
    & \textbf{Hard}
    &
  \\
      $\mDiff(\mArg)
        = \left\{\!
            \begin{array}{ll}
              1                         
                ,&  \!|\mArg| \leq \sqrt{2}\,\theta  \\
              0 
                ,&  \!|\mArg| >    \sqrt{2}\,\theta
            \end{array}
          \right.$     
    & $\mPen(\mArg) 
        = \left\{\!
            \begin{array}{ll}
              \mArg^2                 
                ,&  \!|\mArg| \leq \sqrt{2}\,\theta  \\
              2  \theta^2 
                ,&  \!|\mArg| >    \sqrt{2}\,\theta
            \end{array} 
          \right.$
    & $\mShrink(\mArg) 
        = \left\{\!
            \begin{array}{ll}
              0                       
                ,&  \!|\mArg| \leq \theta            \\
              \mArg 
                ,&  \!|\mArg| >    \theta          
            \end{array}
          \right.$
    & $\mAct(\mArg) 
        = \left\{\!
            \begin{array}{ll}
              \mArg                  
                ,&  \!|\mArg| \leq \sqrt{2}\,\theta  \\
              0 
                ,&  \!|\mArg| >    \sqrt{2}\,\theta
            \end{array}
          \right.$
  \\[3ex] \hline 
\end{tabular}
        \end{table}
      \end{landscape}

    } 

  \subsection{From Variational Models to Nonlinear Diffusion}
    \label{sec_varmethods_to_diffusion}
    \cite{SchW98} consider an energy functional with a quadratic data 
    term and a regulariser $\mPen(\mArg)$:
    \begin{equation}\label{eq_energy_functional}
      E(u) = \int_{\Omega} 
               \left(
                 \left(u - f\right)^2 
                 + \alpha \, \mPen (\dx u)
               \right)
             dx.
    \end{equation}
    The corresponding Euler--Lagrange equation for a 
    minimiser $u$ of the functional reads
    \begin{equation}\label{eq_diffusion_varmethods}
      \frac{u - f}{\alpha} = \dx \left(
                                        \frac{\mPen^\prime (\dx u)}{2}
                                      \right).
    \end{equation}
    This can be regarded as a fully implicit time discretisation for 
    a nonlinear diffusion process with stopping time $T=\alpha$ and 
    diffusivity $\mDiff(\mArg)=\frac{\mPen^\prime(\mArg)}{2\mArg}$. 
    This process can also be approximated by $m$ explicit diffusion 
    steps of type \eqref{eq_nonlinear_diffusion_discrete} with time step 
    size $\tau = \frac{\alpha}{m}$ \cite{RSW99a}, where $m$ is chosen
    such that the stability condition (\ref{eq:stab}) holds.

    A related interpretation which is essentially equivalent to 
    the concept of iterated regularisation is given by algorithm unrolling 
    \cite{KKHP17,SABE18,MLE19}.


  \subsection{Stability Guarantees for Our Residual Network}
    The connections established so far imply direct stability guarantees 
    for networks consisting of diffusion blocks.

    \begin{mytheorem}
      A residual network chaining any number of diffusion blocks with 
      time step size $\tau$, grid size $h$, and antisymmetric activation 
      function $\mAct(\mArg)$ with finite Lipschitz constant $L$ is stable 
      in the $L^\infty$ norm if
      \begin{equation} \label{eq:stab2}
        \tau \leq \frac{h^2}{2L},  
      \end{equation}
      It is also sign stable if the bound is chosen half as large.

    \end{mytheorem}

    Since $\mAct(\mArg) = \mDiff(\mArg) \, \mArg$ and $g$ is a 
    nonincreasing
    symmetric diffusivity with bound $\mDiff_\text{max}$, 
    it follows that $L=\mDiff_\text{max}$. Thus, (\ref{eq:stab2}) is the
    network analogue of the stability condition (\ref{eq:stab}) for an 
    explicit diffusion step. In the same way, the diffusion block inherits
    its sign stability from the sign stability condition of wavelet 
    shrinkage. Stability of the full network follows by induction.  
    
    Note that our results in terms of $L^\infty$ or sign stability are
    stricter stability notions than the $L^2$ stability in \cite{RH20} 
    where more general convolution filters are investigated:
    An $L^2$ stable network can still produce overshoots which violate
    $L^\infty$ stability.

    Contrary to \cite{RH20}, our stability result does not require 
    activation functions to be monotone. We will see that 
    widely used diffusivites and shrinkage functions naturally lead 
    to nonmonotone activation functions.


\section{Dictionary of Activation Functions}\label{sec_discussion}

\subsection{Main Result}

  Exploiting the Equations \eqref{eq_diff_act_connection}, 
  \eqref{eq_diffusion_wavelet}, and \eqref{eq_diffusion_varmethods}, we are 
  now in the position to present a general dictionary which can be used to 
  translate arbitrary diffusivities, wavelet shrinkage functions, and 
  variational regularisers into activation functions. This dictionary is 
  displayed in Table~\ref{table_equivalences}. 
  
  On one hand, our dictionary provides a blueprint for researchers 
  acquainted with diffusion, wavelet shrinkage or regularisation to 
  build a residual network for a desired model while preserving 
  important theoretical properties. This can help them to develop 
  rapid prototypes of the corresponding filters without the need to 
  pay attention to implementational details. Also parallelisation 
  for GPUs is readily available. Last but not least, these methods 
  can be gradually refined by learning. 
  
  On the other hand, also CNN researchers can benefit. The dictionary
  shows how to restrict CNN architectures or parts thereof to models which 
  are well-motivated, provably stable, and can benefit from the rich
  research results for diffusion, wavelet shrinkage and
  regularisation. Lastly, it can inspire CNN practitioners to use 
  more sophisticated activation functions, in particular antisymmetric and 
  nonmonotone ones. 


\subsection{What can We Learn from Popular Methods?}
  Let us now apply our general dictionary to prominent diffusivities,
  shrinkage functions, and regularisers in order to identify their 
  activation functions. 

  We visualise these functions in Table \ref{table_plots} and display
  their mathematical formulas in Table \ref{table_equations}. For our 
  examples, we choose a grid size of $h = 1$. As we have 
  $\mDiff_\text{max} = 1$ for all cases considered, we set
  $\tau = \alpha = \frac14$. This fulfils the sign stability 
  condition (\ref{eq:stab2}).

  We generally observe that all resulting activation functions 
  $\mAct(\mArg) = \mDiff(\mArg) \, \mArg$ are antisymmetric, since 
  $\mDiff(\mArg) = \mDiff(-\mArg)$. This is very natural in the
  diffusion case, where the argument of the flux function is the signal
  derivative $\dx u$. It reflects a desired invariance axiom of 
  denoising: Signal negation and filtering are commutative.
 
  Still, the discussed antisymmetric activation functions can be expressed  
  with typical ReLU functions \cite{NH10}. The truncated total variation 
  activation function
  \begin{equation}
    \mAct\left(\mArg\right) = \left\{\!
                \begin{array}{ll}
                  \mArg                  
                    ,&  \!|\mArg| \leq \sqrt{2}\,\theta  \\
                  \sqrt{2}\,\theta\, \text{sgn}(\mArg)
                    ,&  \!|\mArg| >    \sqrt{2}\,\theta
                \end{array}
              \right.
  \end{equation}
  provides a simple example as it can be rewritten as
  \begin{equation}
  \resizebox{0.85\linewidth}{!}{$
    \mAct\left(\mArg\right) = 
       \mArg 
       - \text{ReLU}\left(\mArg - \sqrt{2}\,\theta \right)
       + \text{ReLU}\left(-\mArg - \sqrt{2}\,\theta\right).
  $}
  \end{equation}
  Other activation functions which are not piecewise linear can be approximated 
  with a series of feedforward layers \cite{HSW89}. By allowing more advanced 
  antisymmetric activation functions, we can end up with fewer layers. It is 
  surprising that this idea has remained basically unexplored in modern CNNs.
    
  Our six examples in Tables \ref{table_plots} and \ref{table_equations} 
  fall into two classes: The first class comprises diffusion filters with 
  constant \cite{Ii62} and Charbonnier diffusivities \cite{CBAB94}, 
  as well as soft wavelet shrinkage \cite{Do95} which involves a Huber 
  regulariser \cite{Hu73} and a truncated total variation (TV) diffusivity 
  \cite{ROF92,ABCM98}. These methods have strictly convex regularisers,
  and their shrinkage functions do not approximate the identity function
  for $\mArg \to \pm \infty$. Most importantly, their activation functions
  are monotonically increasing. This is compatible with the standard scenario
  in deep learning where the ReLU activation function dominates \cite{NH10}. 
  On the diffusion side, the corresponding increasing flux functions act 
  contrast reducing. Strictly convex regularisers have unique 
  minimisers, and popular minimisation algorithms such as gradient 
  descent converge globally.
  
  The second class is much more exciting. Its representatives are given
  by Perona--Malik diffusion \cite{PM90} and two wavelet shinkage methods:
  garrote shrinkage \cite{Ga98} and hard shrinkage \cite{Ma98}. Garrote
  shrinkage corresponds to the truncated balanced forward--backward
  (BFB) diffusivity of \cite{KS02}, while hard shrinkage has a truncated
  quadratic regulariser which is used in the weak string model of
  \cite{GG84}. Approaches of the second class have nonconvex regularisers,
  which may lead to multiple energy minimisers. Their shrinkage functions
  converge to the identity function for $\mArg \to \pm \infty$. The flux
  function of the diffusion filter is nonmonotone. While this was 
  considered somewhat problematic for continuous diffusion PDEs, it has been
  shown that their discretisations are well-posed \cite{WB97}, in spite
  of the fact that they may act contrast enhancing. Since the activation 
  function is equivalent to the flux function, it is also nonmonotone. 
  This is very unusual for CNN architectures. Although there were a few
  very early proposals in the neural network literature arguing that such 
  networks offer a larger storage capacity \cite{FMNP93} and have some 
  optimality properties \cite{MR94}, they had no impact on modern CNNs. 
  
  Our results motivate the idea of nonmonotone activation 
  functions from a different perspective. Since it is well-known that 
  nonconvex variational methods can outperform convex ones, it appears 
  promising to incorporate nonmonotone activations into CNNs in spite of 
  some challenges that have to be mastered.
 

\section{Conclusions}\label{sec_conclusions}

 We have seen that CNNs and classical methods have much more in common 
 than most people would expect: Focusing on three classical denoising 
 approaches in a 1D setting and on a ResNet architecture with simple 
 convolutions, we have established a dictionary that allows to 
 translate diffusivities, shrinkage functions, and regularisers into 
 activation functions. This does not only yield strict stability results 
 for specific ResNets with an arbitrary number of layers, but also 
 suggests to invest more efforts into the design of activation functions. 
 In particular, antisymmetric and nonmonotone activation functions warrant
 more attention. In our current work, we are investigating their usefulness 
 for prototypical classification problems with CNNs.

 Needless to say, our restrictions to 1D, to a single scale, and to
 denoising methods have been introduced mainly for didactic reasons.
 We see our work as an entry ticket to guide also other researchers
 with an expertise on PDEs, wavelets, and variational approaches
 into the CNN universe. As a result, we envision numerous generalisations,
 including extensions to higher dimensions and manifold-valued data. 
 Of course, also additional key features of CNNs deserve to be 
 analysed, for instance pooling operations.
 Our long-term vision is that this line of research will help to bridge
 the performance gap as well as the theory gap between model-driven
 and data-driven approaches, for their mutual benefit.

\section*{Acknowledgements} 
This work has received funding from the European Research Council 
(ERC) under the European Union's Horizon 2020 research and 
innovation programme (grant agreement no. 741215, ERC Advanced 
Grant INCOVID). We thank Michael Ertel for checking our mathematical 
formulas.

\bibliography{myrefs}
\bibliographystyle{icml2020}

\end{document}